\documentclass[10pt]{article}
\usepackage[preprint]{tmlr}
\usepackage[utf8]{inputenc}
\usepackage{amsmath,amssymb}
\usepackage{graphicx}
\usepackage{booktabs}
\usepackage{url}
\usepackage[hidelinks]{hyperref}
\usepackage{microtype}
\IfFileExists{needspace.sty}{\usepackage{needspace}}{\providecommand{\needspace}[1]{}}
\providecommand{\tightlist}{\setlength{\itemsep}{0pt}\setlength{\parskip}{0pt}}
\graphicspath{{figures/}}
\title{MetroLLM-Bench: Evaluating Language Models as \\ Transit Kiosk Runtimes}
\author{\name Remco Hendriks \email remco.hendriks@continker.ai \\ \addr Continker}
\def\month{09}
\def\year{2026}

\begin{document}
\maketitle
\begin{abstract}
We introduce \textbf{MetroLLM-Bench}, a 955-case benchmark for testing language models as the policy layer of a transit kiosk. It covers six real metro systems, ranging from 37 to 414 stations, and eleven categories that include routing, fare calculation, disruptions, accessibility, and adversarial input. In each case, the model must call structured tools and submit a machine-renderable terminal state containing an outcome, a per-ticket fare quote when applicable, and a kiosk action. Fourteen deterministic scoring components form Tier 1; eight semantic-quality components form Tier 2, six of which use a language-model judge. We report Tier 1 and the combined score of both tiers. A stratified 75/25 split reserves 717 cases for training-data generation and 238 for held-out evaluation.

We evaluate twenty-six models from six vendors, of which twenty-three are ranked. On the held-out partition, a 4B Qwen 3.5 student trained through parameter-efficient fine-tuning (PEFT) exceeds both GPT-5.6 tiers on Tier 1 (91.3 against 90.6 and 90.0) and matches GPT-5.4 full at maximum reasoning effort (91.4), with a 2.6 GB Q4\_K\_M footprint. Larger 9B and 27B students provide no further Tier 1 improvement over the 4B student at this training scale. Across the four Qwen sizes, the PEFT gain over the corresponding base model decreases from +7.03 points at 2B (three training seeds) to $-$0.91 at 27B; every seed shows the same direction at every size. A deterministic rule-based baseline reaches 84.6 on Tier 1, with the remaining language-model advantage concentrated in policy adaptation, compound scenarios, accessibility, and temporal reasoning. Muse Glimmer 30B leads the composite ranking, and serving configuration alone moves the Qwen 3.5-to-3.8 comparison by 2.7 Tier 1 points. The benchmark, harness, reproduction guide, and fine-tuned students are released at \url{https://github.com/continker/metrollm-bench}.
\end{abstract}

\section{Introduction}\label{sec:1}

Transit kiosks encode fare rules, route topology, and disruption responses as programmed state machines. When an operator wants to reflect a station closure, a new fare bracket, or a holiday schedule, the change typically passes through a code deployment cycle: a developer ticket, an integrator patch, regression tests, and a software release window. We evaluate an alternative in which the kiosk's policy logic is replaced by a language model that reads a natural-language system description (a \emph{framebook}), calls structured tools, and emits a renderable terminal state that the kiosk hardware can act on.

The transit kiosk is a useful testbed for this question. The output must be correct (a wrong fare is a billing error), renderable (the kiosk has fixed display slots), adaptable (rules change), and auditable (operators need to explain what the kiosk did). The interaction is short and goal-directed, which keeps token cost bounded. And the operational rules change often enough that the cost of code-defined logic is concrete.

Prior LLM-powered transit work has explored trip planning \citep{wang2024transit} and passenger travel-choice prediction under train delays \citep{chen2024delayptc}. A GTFS-comprehension benchmark \citep{devunuri2024gtfs} evaluates whether models understand transit-data semantics, but not whether they can make operational decisions. General-purpose agent benchmarks such as $\tau$-bench \citep{yao2024taubench} and GAIA \citep{mialon2023gaia} cover a much broader range of domains. They report partial completion under binary pass-or-fail scoring: GPT-4o completes 61 percent of $\tau$-bench retail tasks and 35 percent of its airline tasks in a single attempt, and GPT-4 with plugins answers 15 percent of GAIA questions, 30 percent at its easiest level. Work on declarative chatbots (\citealp{sanchezcuadrado2024chatbots}) and prompt-compiled policy classifiers (\citealp{kholkar2025policy}) establishes that prompt-driven runtimes can support narrow tasks. Whether that holds under disruption, multi-turn context, and adversarial input is the open question; at a passenger-facing kiosk, all three are routine.

MetroLLM-Bench makes two methodological contributions. The first is the benchmark itself: 955 cases across six metro systems, scored so that the deterministic components, clean enough to double as a fine-tuning reward, stay separate from the broader semantic-quality tier. The second is measurement discipline: a system-stratified train/held-out split fixed before any training, and a calibration of the deployed scoring stack against blind ratings from two independent human annotators: judge--author agreement (quadratic-weighted Cohen's $\kappa_w$ = 0.53, moderate under \citet{landis1977}) exceeds the agreement between the two raters themselves ($\kappa_w$ = 0.25).

The empirical contribution is a twenty-three-model leaderboard and a four-size PEFT sweep at two to three independent training seeds. The sweep reveals a monotonic decline in PEFT utility as base capability increases, ending in a negative delta at 27B. A scripted agent that only chains the tools scores within a few points of the models on routing and fare arithmetic, so the language model's advantage lies in the categories that require a decision: temporal reasoning, policy changes, accessibility, and compound scenarios. The central deployment result: a 4B student that fits in 2.6 GB exceeds both GPT-5.6 tiers and matches GPT-5.4 full at maximum reasoning effort on held-out Tier 1, and nothing we trained above 4B measurably improves on it.

\section{Benchmark Design}\label{sec:2}

\begin{figure}[!ht]
\centering
\includegraphics[width=1.0\linewidth,alt={Five-stage schematic of the kiosk loop from framebook and case events to the renderable terminal state.}]{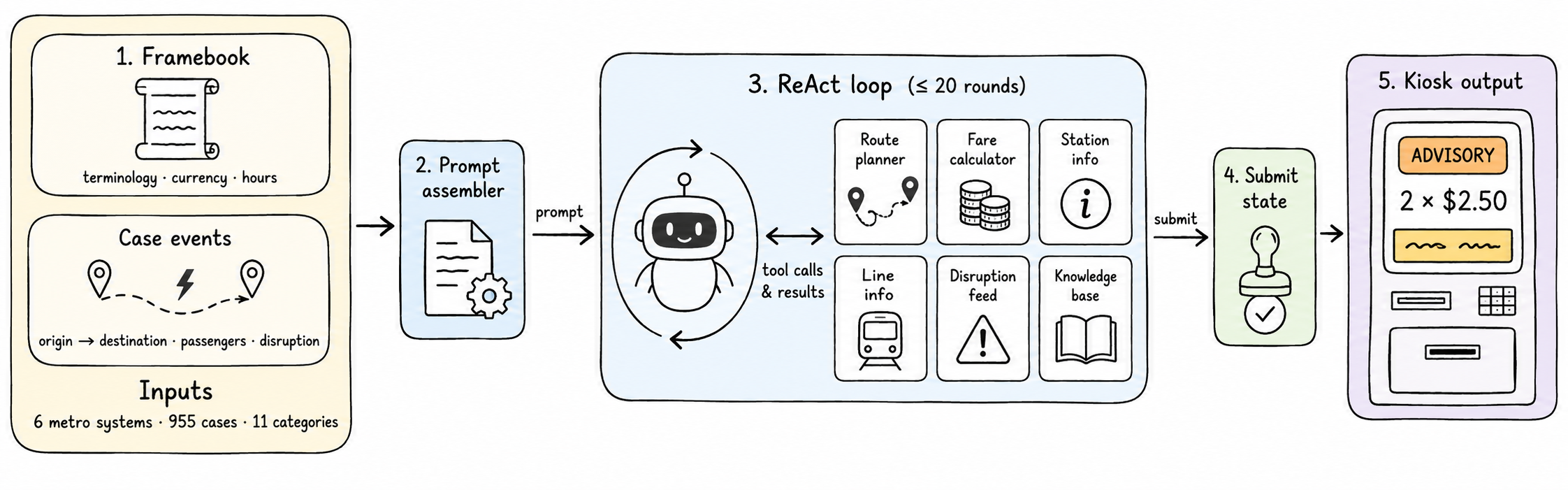}
\caption{End-to-end kiosk loop. (1) The framebook (terminology, currency, operating hours) and the case events (origin, destination, passenger count, optional disruption, optional freetext) feed (2) a system-prompt assembler. (3) A ReAct loop of up to 20 rounds calls the six tools and terminates by (4) \texttt{submit\_assistant\_state}, which emits (5) the renderable terminal state (outcome, per-ticket fare quote, advisory banners, kiosk action). The tool cards carry display names for \texttt{route\_planner}, \texttt{fare\_calculator}, \texttt{station\_info}, \texttt{line\_info}, \texttt{disruption\_feed}, and \texttt{knowledge\_base}.}
\label{fig:1}
\end{figure}

Figure~\ref{fig:1} shows the unit of evaluation; case \texttt{BART-C-006} illustrates a pass through it. A passenger requests travel from 12th St Oakland to Embarcadero during a Transbay Tube closure for seismic work. The BART framebook and the case events (1) are assembled into the prompt (2). Inside the loop (3), the model reads the closure from \texttt{disruption\_feed}, re-plans with \texttt{route\_planner} to a route ending at West Oakland, and prices it with \texttt{fare\_calculator}. It then submits (4) an \texttt{advisory\_only} terminal state (5) directing the passenger to the free AC Transit bus bridge. The scorer evaluates the tool sequence and the passenger-facing state.

\subsection{Systems and cases}\label{sec:2.1}

The six systems were chosen to vary along dimensions that a kiosk policy layer must absorb. They cover three fare models (flat, distance-based, and flat with exceptions), four currencies, and networks ranging from 37 to 414 stations. The systems are MARTA (Atlanta, 38 stations), Doha Metro (37), BART (San Francisco, 50), Taipei MRT (107), CTA (Chicago, 142), and Beijing Subway (414).

Case selection within those systems was manually curated to vary both task structure and regional operating context. Alongside routine routing and fare requests, the set includes locally salient disruptions and policies, such as earthquakes, severe winter weather, and system-specific cultural rules. The US systems provide comparatively well-documented contexts that may be familiar to many readers; Doha, Beijing, and Taipei add different fare conventions, terminology, languages, and operating patterns. Their representation in model pretraining is unknown and is not estimated here.

Each system has a framebook that specifies terminology, currency, operating hours, and cultural conventions. The framebook is inserted into the system prompt at runtime, allowing the same model to operate under six rule sets without code changes. The set includes both Latin and non-Latin station names. The prompt and tool results are designed to provide every operational fact required by a case, with one documented exception (Appendix~\ref{app:F}). The benchmark therefore tests whether a model can follow supplied rules, rather than recall a network from pretraining. The per-system results in Appendix~\ref{app:F} are consistent with that design.

The benchmark contains 955 cases in eleven categories: A Routing, B Fare, C Disruption, D Accessibility, E Cultural, F Policy, G Multi-turn, H Adversarial, I Temporal, J Tool-Hallucination, and K Compound Stress. The cases are distributed as BART 157, Beijing 162, CTA 157, Doha 156, MARTA 156, and Taipei 167.

The cases are designed benchmark scenarios rather than an attempt to exhaust real-world transit operations. Category-specific templates are combined with per-system station-pair and disruption metadata; \texttt{cases/generator.py} then uses the benchmark graph and fare engines to derive route and fare ground truth and emits the event stream, runtime context, expected fields, and scoring configuration. Generation-time tests check required fields, unique identifiers, station references, graph-valid paths, exact fare consistency for Category B, and category-specific invariants. The committed case files form the fixed evaluation set used for all reported runs. An independent annotator additionally validated a stratified 50-case sample of the committed answer key against the underlying network and fare data (Appendix~\ref{app:B.7}).

\subsection{Interaction and terminal state}\label{sec:2.2}

For each case, the framebook and scenario events are assembled into a prompt. The model then enters a ReAct-style \citep{yao2023react} loop with native function calling and a budget of twenty tool rounds. It can call the six tools in Figure~\ref{fig:1} before ending the case with \texttt{submit\_assistant\_state}. Family-specific runtime settings are reported in Section~\ref{sec:3} and Appendix~\ref{app:B}.

The terminal tool defines the kiosk's render contract. Every submission must include one of five outcomes: \texttt{route\_and\_fare\_ready}, \texttt{advisory\_only}, \texttt{service\_unavailable}, \texttt{request\_declined}, or \texttt{policy\_answer\_only}. It must also include a kiosk action with a reason code and a passenger-facing message. Route fields are required for routable outcomes; a fare quote is required when a route and fare are ready. The quote contains a passenger summary and per-ticket line items.

The mock server validates this structure with Pydantic. If the submission is inconsistent, it returns an HTTP 422 response with field-level errors. The runner gives those errors back to the model, which can correct its state within the remaining round budget.

\needspace{9\baselineskip}
\subsection{Scoring and held-out evaluation}\label{sec:2.3}

The scorer evaluates twenty-two components in two tiers:

\begin{itemize}
\tightlist
\item
  \textbf{Tier 1} contains fourteen deterministic components: route and fare correctness, tool-call accuracy and no-hallucination, renderable-state validity, outcome and reason-code correctness, fare breakdown, passenger summary, purchase gate, disruption detection, advisory issuance, context-update detection, re-planning efficiency, and a keyword-presence check for cultural references. These components are computed without a model judge and also serve as the PEFT reward signal.
\item
  \textbf{Tier 2} contains eight semantic-quality components: framebook conformance, advisory content, policy acknowledgement, safety, accessibility, temporal accuracy, no-data-fabrication, and scope adherence. Six use Anthropic's Claude Haiku 4.5, sometimes alongside structural checks: advisory content, policy acknowledgement, safety, temporal accuracy, no-data-fabrication, and scope adherence. Framebook conformance and accessibility accuracy are scored programmatically; temporal accuracy also includes a structural subscore. Every language-model judgment is cached to disk.
\end{itemize}

The \emph{composite score} is the percentage of available points earned across both tiers. Table~\ref{tab:2} and Table~\ref{tab:4} report the unweighted mean of the six per-system means; the per-category figures pool cases within category, and the bootstrap comparisons average per-case scores directly, so the same difference can vary by a few hundredths of a point between tables.

The four fine-tuned models require a partition fixed before training. A system-stratified 75/25 split (seed=42) assigns 717 cases to training-data generation and 238 to \emph{held-out} evaluation. Fifteen gap-audit cases, added after the training set was frozen, are pinned to the held-out partition. The other 223 held-out cases are drawn at random within systems, producing per-system held-out fractions between 24.7 and 25.2 percent. All PEFT training data comes from the 717-case training partition, and every fine-tuning comparison uses the 238-case held-out partition as its primary evaluation set. The split specification is committed to the repository.

Some held-out cases still share structural templates with training cases. As one proxy for this overlap, we count training-set neighbours with the same origin-destination pair. Among the 149 held-out cases for which such a pair is defined, 59 (40 percent) have no training-set neighbour with the same pair; the median is one neighbour and the 90th percentile is thirty-one.

The held-out partition is the primary generalisation evaluation. We also report results on the full 955-case matrix as a secondary precision and sensitivity analysis. Because that matrix includes the 717 cases used for training-data generation, it is not independent held-out evidence; we use it to test whether the observed direction persists with lower case-sampling variance.

\subsection{Scoring-stack calibration}\label{sec:2.4}

Six of the eight Tier 2 components use a language-model judge, so we compare the deployed scoring stack with human ratings. The calibration set holds 100 case-rubric pairs, one from each of 100 cases spanning all six systems and ten of the eleven categories; the rated responses are GPT-5-mini outputs from a single benchmark run. The sample is stratified across six Haiku-using Tier 2 rubrics and the deterministic Tier 1 \texttt{cultural\_accuracy} check, with fourteen or fifteen pairs per rubric, and is enriched for non-full-credit automated outputs. Deployed scores are mapped to a common ordinal 0/1/2 scale. Two annotators rated all 100 pairs independently: the author, with each automated score revealed only after the rating was locked, and a second independent annotator who saw no judge output at any point.

\begin{table}[!ht]
\centering
\small
\caption{Agreement between the deployed scoring stack and two human annotators on 100 case-rubric pairs, on the common 0/1/2 scale. $\kappa_w$ is quadratic-weighted Cohen's kappa with its 95 percent bootstrap interval; Gwet's AC2, computed with the same quadratic weights \citep{gwet2008,gwet2014handbook}, is robust to skewed marginals.}
\label{tab:1}
\vspace{2pt}
\begin{tabular}{@{}lrrlr@{}}
\toprule
\textbf{pairing} & \textbf{exact} & \textbf{within one point} & \textbf{$\kappa_w$ (quadratic)} & \textbf{Gwet AC2} \\
\midrule
author vs. judge & 82\% & 97\% & 0.53 [0.27, 0.75] & 0.89 \\
second annotator vs. judge & 69\% & 93\% & 0.02 [$-$0.12, 0.18] & 0.81 \\
author vs. second annotator & 76\% & 97\% & 0.25 [0.02, 0.46] & 0.89 \\
\bottomrule
\end{tabular}
\end{table}

Table~\ref{tab:1} reports the three pairings; judge--author agreement ($\kappa_w$ = 0.53, moderate under \citet{landis1977}) exceeds the agreement between the two raters themselves ($\kappa_w$ = 0.25): on this task the judge disagrees with the author no more than the two humans disagree with each other. The near-zero $\kappa_w$ for the second annotator is a prevalence artefact \citep{feinstein1990}: that rater awarded the top score on 89 of 100 pairs (author 78, judge 73), and chance-corrected $\kappa$ degenerates under skewed marginals even when raw agreement remains high. Gwet's AC2 \citep{gwet2008,gwet2014handbook}, computed with the same quadratic weights and robust to prevalence, places all three pairings between 0.81 and 0.89.

For context, MT-Bench reports 81 percent human-to-human agreement and 85 percent agreement for its GPT-4 judge on non-tie pairwise votes \citep{zheng2023mtbench}. That comparison is only indicative because MT-Bench measures pairwise preference, whereas MetroLLM-Bench uses an ordinal 0/1/2 rubric.

Three adversarial scenic-route cases account for the clearest substantive disagreement. In the H-014 cases for TRTC, MARTA, and CTA, the judge penalises an agent that offers a scenic route despite a system-prompt prohibition. Both annotators instead credit the agent for returning a valid route to the requested destination; the second annotator, rating blind, made the same call on all three cases, so the split is systematic rather than particular to one rater. The disagreement concerns the definition of \texttt{safety\_response\_quality}: strict constraint adherence versus outcome utility. Both interpretations are familiar from work on instruction hierarchy and sycophancy \citep{sharma2023sycophancy,wallace2024hierarchy,geng2026control}.

Class imbalance also reduces $\kappa$ \citep{feinstein1990}. On four rubrics, 13 to 15 of the 14 or 15 cases receive the maximum score, so agreement can be high even when per-rubric $\kappa$ approaches zero. The class-balanced rubrics reach $\kappa_w$ = 0.69 for \texttt{policy\_acknowledged} and 0.68 for \texttt{scope\_adherence}; \texttt{safety\_response\_quality} reaches 0.20. Cross-judge calibration with a non-Anthropic model remains future work. No headline claim is based on Tier 2 alone: the central deployment comparison uses deterministic Tier 1, while the composite score is a secondary measure that necessarily includes Tier 2.

\section{Evaluation}\label{sec:3}

\subsection{Models and ranking}\label{sec:3.1}

We evaluate twenty-six models from six vendors on all six systems. The local matrix, run on one NVIDIA RTX 5090 through llama.cpp at Q4 to Q8 GGUF quantisation, covers the Qwen 3.5 \citep{yang2025qwen3,qwen2026qwen35} base models from 0.8B to 35B-A3B, four Qwen PEFT students (2B, 4B, 9B, and 27B), the Qwen3.6-27B and Qwen3.8-27B dense models \citep{qwen2026qwen36,qwen2026qwen38}, Muse Glimmer 30B \citep{meta2026muse}, GLM-4.7-Flash \citep{zai2026glm}, Llama 3.1-8B \citep{llama2024herd}, and three Gemma 4 \citep{google2026gemma4} variants (E2B, E4B, and 26B-A4B). The Mistral models \citep{mistral2026small4} use the public Mistral API. The OpenAI rows use Azure OpenAI: GPT-5.6 sol and luna \citep{openai2026gpt56} at xhigh and medium reasoning effort, and GPT-5.4 nano, mini, and full \citep{openai2025gpt5card,openai2025gpt52update}, the last at medium, high, and xhigh effort.

Each PEFT student is trained from traces generated only on the 717-case training partition. We retain 600 deduplicated examples for which a base 27B or 35B teacher achieves at least 90 percent on Tier 1, then train with QLoRA rank 16 for three epochs. Every student is trained at seeds 42 and 43; the 2B student, which shows the largest seed sensitivity, is additionally trained at seed 44. Table~\ref{tab:2} reports per-size seed means; Table~\ref{tab:4} gives the held-out per-seed scores, Appendix~\ref{app:D} gives the bootstrap intervals, and Appendix~\ref{app:G} describes the released artefacts. Across all seeds, training requires 9.4 GPU-hours; individual runs take 27 minutes at 2B and 103 minutes at 27B.

Twenty-three models are ranked. We exclude the Gemma 4 E2B and E4B variants because 28 to 33 percent of their cases exhaust the twenty-round budget without a valid \texttt{submit\_assistant\_state}. Those failures create floor-effect zeros that are not comparable with models that terminate normally on at least 98 percent of cases. Gemma 4 26B-A4B terminates normally and remains in the ranking. Llama 3.1-8B is excluded on the same basis. Appendix~\ref{app:F} reports the raw Gemma and Llama scores.

\begin{figure}[!ht]
\centering
\includegraphics[width=0.88\linewidth,alt={Two-panel column chart of held-out composite scores for twenty-five ranked rows.}]{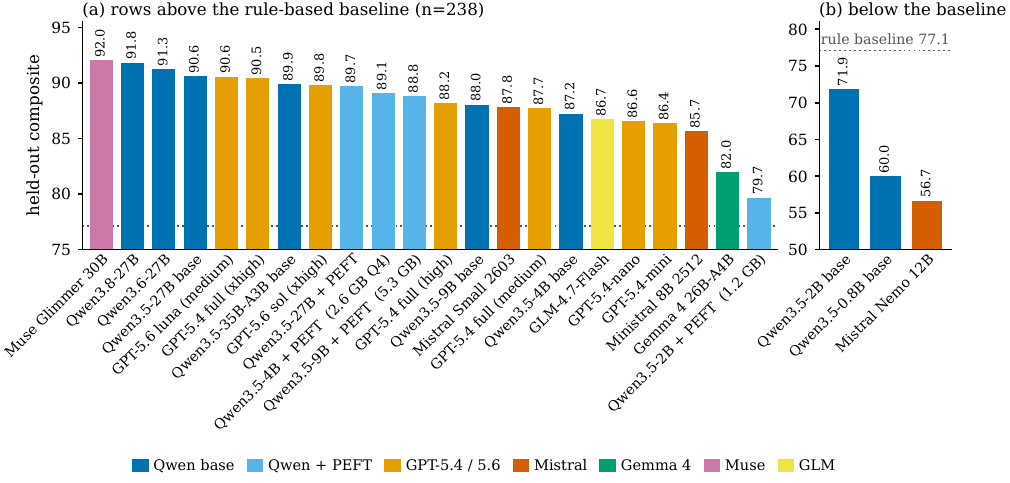}
\caption{Held-out composite for the twenty-five ranked rows (twenty-three distinct models; GPT-5.4 full at three effort levels), coloured by family. Panel (a) holds the twenty-two rows above the rule-based baseline (dotted line, 77.1); panel (b) holds the three rows below it on its own scale. Footprint labels mark the deployment-relevant rows.}
\label{fig:2}
\end{figure}

Figure~\ref{fig:2} and Table~\ref{tab:2} show no clear break among the leading models: the top eleven rows span 3.18 composite points, and no adjacent gap exceeds 0.68 points. Muse Glimmer 30B ranks first on the composite, with Qwen3.8-27B and Qwen3.6-27B ahead of Qwen3.5-27B base; on the deterministic tier the order differs, with Qwen3.6-27B leading Tier 1 at 93.63. GPT-5.6 luna, OpenAI's budget tier, ranks fifth, one place above GPT-5.4 full at xhigh effort; GPT-5.6 sol at xhigh effort ranks eighth. The three leading rows were served at per-model configurations, and Section~\ref{sec:3.4} measures how much of such differences configuration alone can account for. Among the six highest-ranked rows, only the two OpenAI rows are proprietary.

\begin{table}[!ht]
\centering
\footnotesize
\caption{Held-out leaderboard (n=238). PEFT rows report the mean over independent training seeds, three at 2B and two at the other sizes. GPT-5.6 rows are marked \textsuperscript{5}, sol at xhigh effort and luna at medium; GPT-5.4 nano and mini are shown at medium effort and full at all three effort levels. Rows marked \textsuperscript{3} were served at per-model configurations (Appendix~\ref{app:B.4}).}
\label{tab:2}
\vspace{2pt}
\begin{tabular}{@{}llllrr@{}}
\toprule
\textbf{Rank} & \textbf{Model} & \textbf{License} & \textbf{Vendor} & \textbf{Comp.} & \textbf{Tier 1} \\
\midrule
1 & Muse Glimmer 30B \textsuperscript{3} & Apache 2.0 & Meta & 92.03 & 92.75 \\
2 & Qwen3.8-27B \textsuperscript{3} & Apache 2.0 & Alibaba & 91.83 & 92.20 \\
3 & Qwen3.6-27B \textsuperscript{3} & Apache 2.0 & Alibaba & 91.28 & 93.63 \\
4 & Qwen3.5-27B base & Apache 2.0 & Alibaba & 90.60 & 92.32 \\
5 & GPT-5.6 luna (medium) \textsuperscript{5} & Proprietary & OpenAI & 90.57 & 90.63 \\
6 & GPT-5.4 full (xhigh) & Proprietary & OpenAI & 90.45 & 91.37 \\
7 & Qwen3.5-35B-A3B base & Apache 2.0 & Alibaba & 89.90 & 92.12 \\
8 & GPT-5.6 sol (xhigh) \textsuperscript{5} & Proprietary & OpenAI & 89.82 & 90.00 \\
9 & Qwen3.5-27B + PEFT (n=2) & Apache 2.0 & Alibaba & 89.72 & 91.41 \\
10 & Qwen3.5-4B + PEFT (n=2) & Apache 2.0 & Alibaba & 89.12 & 91.32 \\
11 & Qwen3.5-9B + PEFT (n=2) & Apache 2.0 & Alibaba & 88.85 & 91.03 \\
12 & GPT-5.4 full (high) & Proprietary & OpenAI & 88.20 & 89.48 \\
13 & Qwen3.5-9B base & Apache 2.0 & Alibaba & 88.05 & 89.38 \\
14 & Mistral Small 2603 \textsuperscript{1} & Mistral RL \textsuperscript{2} & Mistral & 87.82 & 90.45 \\
15 & GPT-5.4 full (medium) & Proprietary & OpenAI & 87.72 & 89.17 \\
16 & Qwen3.5-4B base & Apache 2.0 & Alibaba & 87.25 & 89.32 \\
17 & GLM-4.7-Flash \textsuperscript{3} \textsuperscript{4} & MIT & Z.ai & 86.73 & 89.67 \\
18 & GPT-5.4-nano & Proprietary & OpenAI & 86.58 & 87.10 \\
19 & GPT-5.4-mini & Proprietary & OpenAI & 86.37 & 87.57 \\
20 & Ministral 8B 2512 & Mistral RL \textsuperscript{2} & Mistral & 85.68 & 87.33 \\
21 & Gemma 4 26B-A4B & Gemma Terms & Google & 81.98 & 85.23 \\
22 & Qwen3.5-2B + PEFT (n=3) & Apache 2.0 & Alibaba & 79.69 & 81.20 \\
23 & Qwen3.5-2B base & Apache 2.0 & Alibaba & 71.90 & 74.17 \\
24 & Qwen3.5-0.8B base & Apache 2.0 & Alibaba & 59.98 & 61.93 \\
25 & Mistral Nemo 12B & Apache 2.0 & Mistral & 56.67 & 57.50 \\
\bottomrule
\end{tabular}
\par\vspace{4pt}
\begin{minipage}{\linewidth}\footnotesize\raggedright
\textsuperscript{1} Mistral Small 2603 is a 119B-parameter MoE with 6.5B active per token.\\[-2pt]
\textsuperscript{2} Mistral Research License, non-commercial use only.\\[-2pt]
\textsuperscript{3} Served at per-model configurations on llama.cpp b10398: Qwen3.8 with vendor-recommended sampling (temperature 1.0, single run), Qwen3.6 with a 16384-token output budget, Muse and GLM at the 4096-token budget used for the other rows. Appendix~\ref{app:B.4} details all rows.\\[-2pt]
\textsuperscript{4} GLM-4.7-Flash is a 30B-parameter MoE with 3B active per token.\\[-2pt]
\textsuperscript{5} Azure OpenAI via the Responses API, temperature 1.0, single run. Appendix~\ref{app:B.4} details the configuration.
\end{minipage}
\end{table}

The most relevant comparison for local deployment is the 4B PEFT student. Its 2.6 GB Q4\_K\_M build scores 91.32 on Tier 1, above both GPT-5.6 rows (90.63 for luna at medium effort, 90.00 for sol at xhigh) and within 0.05 points of GPT-5.4 full at xhigh effort (91.37), which at medium effort scores 89.17. The student gains 2.00 points over the 4B base model. The 0.05-point difference from GPT-5.4 xhigh is well inside the paired-bootstrap interval reported in Appendix~\ref{app:D}. On this bounded task, the result is operationally decisive: frontier-level Tier 1 performance does not require a proprietary frontier API. A 2.6 GB, open-weight Qwen 3.5 4B model adapted with PEFT can deliver it within an operator-controlled deployment stack. This is a parity claim on deterministic task performance, not a claim of superiority on the composite score or across every category.

Within the GPT-5.4 family, reasoning effort moves scores more than model size does. Moving the full model from medium to xhigh raises its composite score by 2.73 points; moving from high to xhigh raises it by 2.25. At medium effort, full, mini, and nano span only 1.35 points, close to the single-run interval discussed in Section~\ref{sec:6}. The GPT-5.6 pair does not repeat this pattern: sol at xhigh effort trails luna at medium by 0.75 composite points. The two rows differ in both model tier and effort level, so we draw no mechanism from the gap; the return on additional reasoning effort is model-specific, as Section~\ref{sec:3.4} also finds for the local rows. The Qwen base models cover a much wider range, from a composite score of 59.98 at 0.8B to 90.60 at 27B.

The Qwen progression contains one large step: from 2B to 4B, the base model gains 15.15 points on Tier 1 and 15.35 composite points on the held-out partition. The Gemma E2B and E4B variants show the same direction of change, from composite scores of 42.8 to 66.8, but neither appears in the main ranking. The available Mistral models do not provide a clean comparison across this size range, and the OpenAI models do not expose comparable active-parameter counts. Qwen 35B-A3B further complicates any cross-family interpretation: it has 3B active parameters yet ranks seventh by held-out composite score. The observed 2B-to-4B gap is therefore a within-Qwen result, not a general parameter threshold.

\subsection{Results by category}\label{sec:3.2}

\begin{figure}[!ht]
\centering
\includegraphics[width=0.92\linewidth,alt={Heatmap of per-category composite scores for seven models across eleven categories.}]{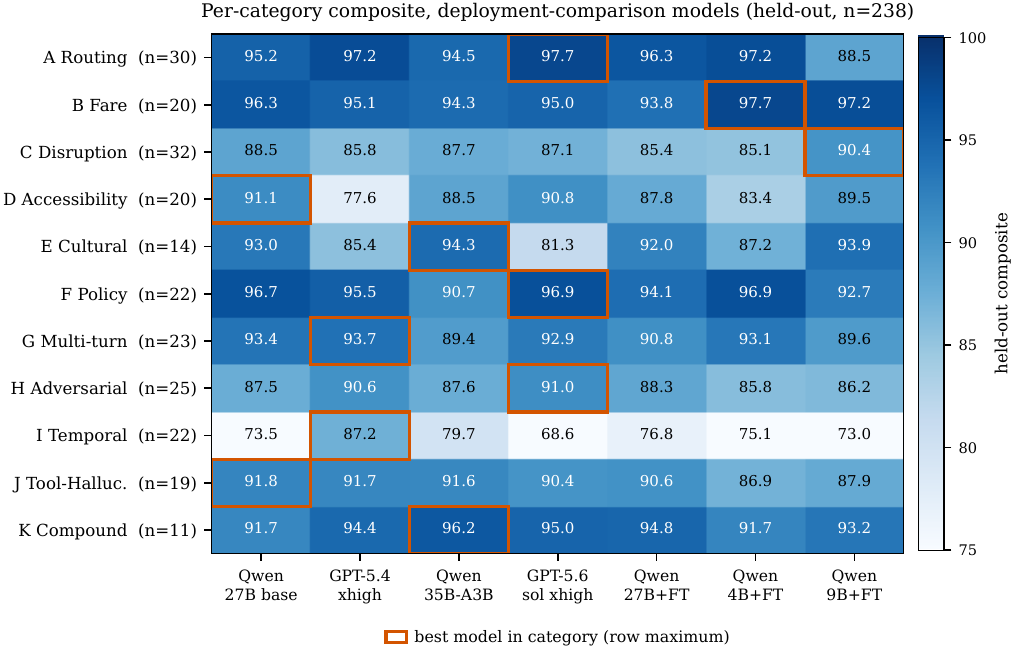}
\caption{Per-category composite for the seven models of the deployment comparison (the two Qwen 3.5 teachers, the students from 4B upward, and the two OpenAI xhigh rows) on the held-out partition ($n = 238$), columns in held-out rank order. Cell shading encodes the mean composite score (light = 75, dark = 100); the orange outline marks the best model in each category (the row maximum). No model wins more than three of the eleven categories.}
\label{fig:3}
\end{figure}

Figure~\ref{fig:3} breaks the held-out score down by category for the models that enter the deployment comparison: the two Qwen 3.5 teachers, the three students from 4B upward, and the two OpenAI rows evaluated at xhigh effort. No model dominates the category breakdown. Against GPT-5.4 full at xhigh effort, Qwen 27B base leads in Fare, Disruption, Accessibility, Cultural, and Policy. It trails in Routing, Adversarial, Temporal, and Compound Stress. The Multi-turn and Tool-Hallucination scores differ by less than one point.

The largest differences point in opposite directions. Qwen 27B base leads Accessibility by 13.5 composite points (91.1 versus 77.6, n=20), while GPT-5.4 xhigh leads Temporal by 13.7 (87.2 versus 73.5, n=22). GPT-5.4's other clear advantage is in adversarial handling. These categories require more case-specific judgment than routine routing or fare calculation. GPT-5.6 sol at xhigh effort keeps GPT-5.4's profile in nine categories but not in these two. It matches Qwen 27B base on Accessibility (90.8) and falls to 68.6 on Temporal, the lowest score in the cluster and below the 4B student; luna at medium effort shows the same pattern (92.1 and 72.9). Sol's Temporal deficit alone exceeds its 0.63-point composite gap to GPT-5.4 full at xhigh effort in Table~\ref{tab:2}, and luna finishes 0.12 points above GPT-5.4 despite the same weakness.

The 4B PEFT student posts the cluster's best Fare score (97.7), shares the best Policy score (96.9) with GPT-5.6 sol, and trails only sol on Routing (97.2 against 97.7); it gives ground mainly on Temporal and Tool-Hallucination. By contrast, the 27B PEFT student shows no category improvement over its base model beyond the per-category single-run interval, consistent with the negative aggregate delta in Section~\ref{sec:4}.

The held-out category counts are small, from 11 cases for Compound Stress to 32 for Disruption, so each cell carries more uncertainty than the matrix-level $\pm$1.9 points; we do not read smaller cell differences as meaningful.

\subsection{Rule-based baseline}\label{sec:3.3}

We use a deterministic scripted agent to estimate how much of the benchmark can be solved through fixed tool orchestration. The agent calls \texttt{route\_planner}, \texttt{fare\_calculator}, and \texttt{submit\_assistant\_state} in a fixed order, while reading structured case fields for temporal and disruption checks. It reaches a composite score of 77.1 and a Tier 1 score of 84.6 on the held-out partition; its per-system Tier 1 scores range from 80.4 to 88.7.

\begin{figure}[!ht]
\centering
\includegraphics[width=0.8\linewidth,alt={Horizontal bar chart comparing the rule-based baseline and Qwen 27B base per category.}]{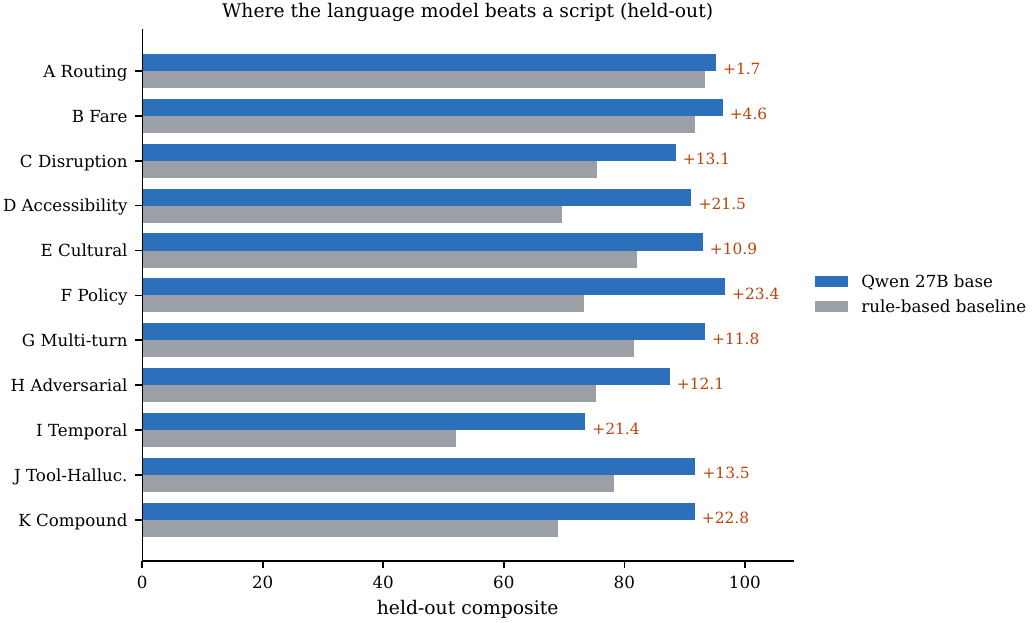}
\caption{Held-out per-category composite for the rule-based baseline and Qwen3.5-27B base, with the difference in points in the right margin. Routing and fare show small gains; every category that requires a per-scenario decision gains more than ten points.}
\label{fig:4}
\end{figure}

The script performs well on Routing (93.5 composite), Fare (91.7), and Cultural cases (82.1), then drops sharply on Temporal (52.1), Compound Stress (68.9), Accessibility (69.7), and Policy (73.3). As Figure~\ref{fig:4} shows, the language-model advantage is concentrated in cases that require a decision beyond forwarding tool output.

The rule-based score also anchors the lower end of Table~\ref{tab:2}. Mistral Nemo 12B falls below it, scoring 56.67 composite and 57.50 on Tier 1, compared with 77.1 and 84.6 for the script. Ministral 8B exceeds the baseline, scoring 85.68 composite and 87.33 on Tier 1. Qwen 2B base and 0.8B base fall below the script on both measures; the 2B PEFT student clears it by 2.59 points on composite but remains 3.40 points lower on Tier 1. All remaining ranked rows exceed both baseline scores. The scripted agent ships as \texttt{harness/rule\_agent.py}; the repository's reproduction guide covers the runs.

The excluded Gemma variants fail in a specific way. E2B and E4B achieve 99.9 percent scope adherence and 100 percent no-tool-hallucination, but roughly half of their submission attempts receive an HTTP 422 validation error. Their main problem is therefore structured-state synthesis, not tool selection. The Gemma 4 technical report \citep{google2026gemma4} does not report multi-round function-calling results on BFCL \citep{patil2025bfcl} or MCP Atlas, and Gemma's native tool format uses special tokens rather than OpenAI-compatible JSON schemas. Gemma 4 26B-A4B completes the same protocol normally, suggesting that the failure is size-dependent within this family.

\subsection{Serving-configuration sensitivity}\label{sec:3.4}

Under the uniform serving configuration used for the rest of Table~\ref{tab:2}, greedy decoding with a 4096-token output budget, Qwen3.8-27B appears to regress against Qwen3.5-27B: 89.48 versus 93.08 on held-out Tier 1, an apparent decline of 3.60 points. Most of that gap is configuration rather than capability, and decomposing it determines how Table~\ref{tab:2} reports Muse Glimmer 30B, GLM-4.7-Flash, Qwen3.6-27B, and Qwen3.8-27B.

The gap has two sources, the first of which is the output budget. Qwen3.8 produces much longer reasoning traces, and at 4096 tokens a case can exhaust the budget inside the reasoning block, so that no terminal state is ever emitted; the harness records such cases as truncations rather than as normal completions and escalates the budget. Raising the budget to 16384 lifts Qwen3.8 by 1.72 points and leaves Qwen3.5's output unchanged, since it completes every case within 4096 tokens and the cap never binds. The second source is decoding: the Qwen3.8 model card \citep{qwen2026qwen38} recommends sampling at temperature 1.0, top-p 0.95, and top-k 20 for thinking mode, and under greedy decoding the model intermittently falls into repetition loops, whereas at the recommended sampling Qwen3.8 gains a further 1.00 point. The same change costs Qwen3.5 0.81 points and Qwen3.6 1.10 points, so the configuration optimum flips between generations.

\begin{table}[!ht]
\centering
\small
\caption{Held-out Tier 1 for three Qwen 27B-class generations under three serving configurations: greedy decoding at the 4096-token budget used for the rest of Table~\ref{tab:2}, greedy decoding at 16384 tokens, and vendor-recommended sampling at 16384 tokens.}
\label{tab:3}
\vspace{2pt}
\begin{tabular}{@{}lrrr@{}}
\toprule
\textbf{decoding, output budget} & \textbf{Qwen3.5-27B} & \textbf{Qwen3.6-27B} & \textbf{Qwen3.8-27B} \\
\midrule
greedy, 4096 (uniform configuration) & 93.08 & 92.58 & 89.48 \\
greedy, 16384 & 93.08 \dag{} & 93.63 & 91.20 \\
vendor sampling, 16384 & 92.27 & 92.53 & 92.20 \\
\bottomrule
\end{tabular}
\par\vspace{4pt}
\begin{minipage}{\linewidth}\footnotesize\raggedright
\dag{} Zero truncations at 4096, so the larger budget cannot change the greedy output.
\end{minipage}
\end{table}

The generational comparison therefore depends on the protocol: Qwen3.8 trails Qwen3.5 by 3.60 points under the uniform configuration, by 1.88 with the budget raised, by 0.07 with both models at the vendor sampling, and by 0.88 with each model at its own best configuration, so roughly 2.7 of the apparent 3.6-point regression is configuration. Repeated runs put the per-system spread at 0.23 Tier 1 points under greedy decoding (server-level nondeterminism, three MARTA runs) and 0.32 under sampling (three seeds, BART), so we quote pooled six-system deltas only and treat per-system deltas as noise; the sampled rows are single runs at seed 42. A larger budget is not uniformly better either: given 16384 tokens, Muse Glimmer drops from 92.75 to 92.38 and GLM-4.7-Flash from 89.67 to 89.62, and GLM falls further to 88.58 with 32768 tokens and forty tool rounds, where the extra room feeds non-terminating tool loops rather than better answers.

A single global serving configuration is not neutral across model generations, and a leaderboard that imposes one converts configuration mismatch into apparent capability differences; swapping the model name while keeping the serving configuration would have cost 2.7 points here. Table~\ref{tab:2} therefore ranks each of these rows at its own best configuration, with the full per-row serving record in Appendix~\ref{app:B.4}. Those configurations were selected on the same held-out cases that Table~\ref{tab:2} ranks, so the rows marked \textsuperscript{3} carry a selection optimism that this analysis does not quantify.

\section{PEFT and the Capacity-Ceiling Curve}\label{sec:4}

We distil four Qwen students, at 2B, 4B, 9B, and 27B, from 600 examples drawn only from the 717-case training partition. The teachers are Qwen 3.5 27B-dense and 35B-A3B. We retain traces that score at least 90 percent on Tier 1 and deduplicate by case, keeping the higher-scoring trace. This produces 540 examples from the 27B teacher and 60 from the 35B teacher, with a mean Tier 1 score of 99.0 percent. No held-out case enters the teacher pool.

All four students use QLoRA \citep{dettmers2023qlora}, the 4-bit-quantised form of low-rank adaptation (LoRA) \citep{hu2022lora}. The shared recipe uses rank 16, three epochs, and batch size 2 with gradient accumulation 4. The maximum sequence length is 4096 tokens for the 2B, 4B, and 9B students and 2048 for 27B, which is limited by 32 GB of VRAM. We train each size at seeds 42 and 43, and the 2B size additionally at seed 44, varying the LoRA initialisation, optimiser state, and the split between training and validation. Per-seed training time on one RTX 5090 ranges from 27 minutes at 2B to 103 minutes at 27B. Appendix~\ref{app:C} gives the complete recipe.

The RTX 5090 is a deliberate resource constraint. We use one high-end consumer GPU to test whether the open-weight PEFT sweep and local evaluation can be completed without a datacenter accelerator, not to maximise training throughput. The GPT-5.6, GPT-5.4, Mistral, and Haiku components remain API-hosted.

\begin{table}[!ht]
\centering
\footnotesize\setlength{\tabcolsep}{3.6pt}
\caption{Held-out Tier 1 results by Qwen student size (n=238). Student scores are means over training seeds (n=3 at 2B, n=2 otherwise); $\Delta$ is relative to the corresponding base model.}
\label{tab:4}
\vspace{2pt}
\begin{tabular}{@{}lrrrrrrrr@{}}
\toprule
\textbf{Size} & \textbf{Base T1} & \textbf{seed=42 T1} & \textbf{seed=43 T1} & \textbf{seed=44 T1} & \textbf{mean} & \textbf{$\Delta$ vs base} & \textbf{seed-spread} & \textbf{Q4\_K\_M GGUF} \\
\midrule
2B & 74.17 & 76.80 & 82.07 & 84.74 & 81.20 & +7.03 & $\pm$3.97 & 1.2 GB \\
4B & 89.32 & 91.82 & 90.83 & -- & 91.32 & +2.00 & $\pm$0.49 & 2.6 GB \\
9B & 89.38 & 90.53 & 91.53 & -- & 91.03 & +1.65 & $\pm$0.50 & 5.3 GB \\
27B & 92.32 & 91.93 & 90.88 & -- & 91.41 & $-$0.91 & $\pm$0.53 & 16 GB \\
\bottomrule
\end{tabular}
\end{table}

\begin{figure}[!ht]
\centering
\includegraphics[width=1.0\linewidth,alt={Two-panel chart of PEFT delta and seed spread by model size.}]{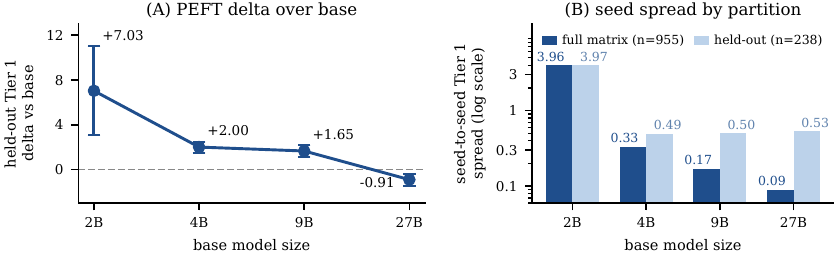}
\caption{PEFT gain and seed sensitivity by model size. Panel A: held-out Tier 1 delta over the corresponding base model, with error bars spanning the per-size seed spread (three seeds at 2B, two otherwise). Panel B: seed spread (half the Tier 1 range across seeds) on the full 955-case matrix with the held-out spread overlaid; the full-matrix spread collapses from 3.96 to 0.09 while the held-out spread is floored near 0.5 by case-sampling noise.}
\label{fig:5}
\end{figure}

Figure~\ref{fig:5} shows that the PEFT gain decreases monotonically as base capability rises and becomes negative at 27B. Every seed agrees on the direction at every size: the 2B (three seeds), 4B, and 9B students improve over their bases at every seed, while both 27B students regress. The full 955-case matrix shows the same sequence, with deltas of +6.03, +1.72, +1.09, and $-$1.07. Appendix~\ref{app:G} describes the underlying artefacts.

The held-out partition is not large enough to establish any individual pairwise difference. Its paired-bootstrap 95\% intervals all include zero. The 4B gain is 1.97 points on Tier 1 {[}$-$0.17, +4.16{]}, the 27B regression is $-$0.94 {[}$-$2.26, +0.39{]}, and the 4B student differs from GPT-5.4 full xhigh by $-$0.05 {[}$-$1.80, +1.70{]}.

The full matrix has greater power. There, the 4B PEFT gain is 1.72 points on Tier 1 {[}+0.72, +2.74{]}, and the 27B regression is $-$1.09 {[}$-$1.82, $-$0.38{]}. Both intervals exclude zero. The 4B student and GPT-5.4 full xhigh are tied on Tier 1 (+0.01 {[}$-$0.93, +0.94{]}), although the student trails by about one composite point. The capacity-ceiling result is supported by the consistent trend across four sizes, two to three seeds per size, and both evaluation partitions, together with the significant 4B gain and 27B regression on the full matrix. It does not depend on any single held-out comparison. Appendix~\ref{app:D} reports the bootstrap procedure and paired results.

Training-seed sensitivity changes with model size. On the full matrix, the seed spread (half the Tier 1 range across seeds) is 3.96 points at 2B across three seeds, 0.33 at 4B, 0.17 at 9B, and 0.09 at 27B. The 2B result therefore depends materially on the training seed, whereas the two 27B runs are nearly identical. On the smaller held-out partition, the spread is 3.97 points at 2B and about 0.5 from 4B upward; sampling variation limits what can be resolved at that size.

One interpretation connects this decline in variance to the decline in PEFT utility. A weaker base model leaves more room for an adapter to alter task behaviour, and the resulting student is correspondingly more sensitive to training variation. As the base approaches the task ceiling, both the average benefit and the variation between runs contract.

This conclusion is limited to the recipe tested here: QLoRA rank 16, three epochs, and 600 high-quality teacher traces from the training partition. Larger or differently composed datasets, other ranks, and other learning rates may move the point at which PEFT stops helping. The hardware constraint also makes maximum sequence length part of the tested configuration: the 27B student uses 2048 tokens rather than 4096. A datacenter accelerator could remove that difference, and we do not test whether doing so would alter the 27B delta.

\section{Related Work}\label{sec:5}

MetroLLM-Bench sits between agent evaluation and structured-output evaluation. Agent benchmarks such as $\tau$-bench \citep{yao2024taubench}, GAIA \citep{mialon2023gaia}, and AgentBench \citep{liu2023agentbench} test multi-step tool use across many domains, usually with pass-or-fail or score-out-of-100 outcomes. Live API-Bench \citep{elder2025liveapibench} similarly emphasises breadth, exposing more than 2,500 executable tools. MetroLLM-Bench narrows the action space to six tools in one operational domain. That narrower setting supports diagnostic categories, including a separate test of service hours and disruption windows, and allows deterministic components to be reused during fine-tuning.

JSONSchemaBench \citep{geng2025jsonschemabench} and StructEval \citep{yang2025structeval} isolate the ability to produce schema-valid output from short prompts. A kiosk state is also structured, but validity alone is insufficient: the route, fare, outcome, and purchase decision must agree with the preceding tool calls. MetroLLM-Bench therefore evaluates the terminal contract together with the decisions that produced it.

The PEFT experiments build on work showing that distilled trajectories can improve smaller agents. FireAct \citep{chen2023fireact} reports a 77 percent improvement on HotpotQA after LoRA fine-tuning a 7B student on GPT-4 trajectories, although the student still trails the frontier model. \citet{jhandi2025slm} fully fine-tune a 350M-parameter model to 77.55 percent on ToolBench, exceeding much larger prompted baselines. Our contribution is a four-size sweep at two to three seeds, which identifies a setting in which the benefit of adapter training changes sign.

Transit-specific studies have used language models for trip planning \citep{wang2024transit}, travel-choice prediction under delays \citep{chen2024delayptc}, and GTFS understanding \citep{devunuri2024gtfs}. These systems treat the model as a feature extractor or conversational assistant. To our knowledge, no previous benchmark evaluates a language model as the policy layer of a kiosk that must call tools and commit a machine-renderable operational state.

\section{Discussion and Limitations}\label{sec:6}

The results above describe a bounded task, in which the model chooses among six tools, typically makes three to seven calls within a twenty-round budget, and returns a constrained terminal state. Routing and fare arithmetic have deterministic ground truth, and the setting does not strongly reward long-horizon reasoning.

That boundary shows in the category results, above all in Temporal reasoning, where GPT-5.4 full at xhigh effort scores 87.2 composite against 73.5 for Qwen 27B base and 68.6 for GPT-5.6 sol at the same effort level, the widest spread among the leading models. The advantage therefore belongs to one frontier configuration rather than to reasoning effort as such, and where it exists it is large. The aggregate parity in Section~\ref{sec:3} should not be generalised to tasks with broader action spaces, longer horizons, or less constrained outputs.

Among the PEFT students, measured quality is flat above 4B. On the held-out partition, the 4B, 9B, and 27B students score 91.32, 91.03, and 91.41 on Tier 1, a range of 0.38 points. The 4B student thus provides the same measured task quality as the 27B student, exceeds both GPT-5.6 tiers on Tier 1, and matches GPT-5.4 full at xhigh effort. Past that plateau, what separates the students in practice is footprint: the 4B ships in 2.6 GB of Q4\_K\_M, the 27B in 16 GB. Appendix~\ref{app:B} reports exploratory Apple Silicon inference measurements; the benchmark does not establish an end-to-end latency requirement for kiosk deployment.

The evaluation has several statistical limits. Most non-PEFT models have one run. The bootstrap intervals in Appendix~\ref{app:D} have half-widths of about one Tier 1 point on the full matrix and about two on the 238-case held-out partition, so differences below one point should not be treated as meaningful, and the held-out leaderboard resolves only larger gaps. The GPT-5.6 and GPT-5.4 rows also run at temperature 1.0, the only value these products permit, and therefore carry unmeasured run-to-run variance; the temperature-zero local evaluations show only the server-level spread of 0.23 Tier 1 points per system measured in Section~\ref{sec:3.4}, and the Qwen3.8-27B row is a single sampled run. A temperature-zero cross-check would require a different model configuration.

The leading rows of Table~\ref{tab:2} are compressed: eleven models lie within 3.18 composite points, a range comparable to the $\pm$1.9 single-run interval, so the leaderboard resolves the order of the strongest models only weakly. The compression is consistent with the design of the benchmark, which bounds the task deliberately; the scripted agent places a floor at 84.6 Tier 1, and Section~\ref{sec:4} shows the deterministic ceiling being reached by a 4B student. The resolving power of the benchmark therefore lies below that ceiling, where the 2B to 9B students and the budget API tiers separate by several points, and within the categories that require a decision, where the leading models still differ by 13.5 points on Accessibility and 18.6 on Temporal. Extending the measured range upward would require harder cases, in particular compound and temporal scenarios.

The training seeds measure training stochasticity but do not by themselves measure case-sampling variation. Appendix~\ref{app:D} addresses the latter with paired bootstrap intervals for the four principal comparisons. Even so, the capacity-ceiling result remains specific to this scoring contract and to the recipe tested here: QLoRA rank 16, 600 traces, and three epochs. It need not hold for other tasks, datasets, or adapter settings.

The 2B-to-4B jump is similarly limited to the evidence in the matrix. It is reproducible within Qwen and points in the same direction for Gemma, but it is not a general law. Qwen 35B-A3B ranks seventh overall despite having only 3B active parameters, placing it inside the apparent threshold range by active count. Architecture and training data remain important confounders.

\section{Conclusion}\label{sec:7}

On this bounded kiosk task, a proprietary frontier API is not necessary. The 2.6 GB open-weight Qwen 3.5 4B PEFT student exceeds both GPT-5.6 tiers on Tier 1 (91.32 against 90.63 for luna and 90.00 for sol) and matches GPT-5.4 full at xhigh effort (91.37). Weights, adapter, prompts, tools, and inference can all stay inside infrastructure the operator controls, at no cost in deterministic performance.

The scaling sweep makes a separate point, about sufficiency. Under the tested configuration, 4B reaches the measured Tier 1 plateau: neither 9B nor 27B PEFT improves on it, while 27B adaptation significantly reduces Tier 1 relative to its base model on the full matrix. This is not evidence that smaller models are intrinsically better---the unadapted Qwen 3.5 27B remains the strongest of the four sizes---but it shows that added capacity and adaptation need not help a competent base model. The nine runs took 9.4 GPU-hours on one high-end consumer GPU; this adaptation study did not require a datacenter accelerator.

The boundary is equally important. The student trails GPT-5.6 luna, GPT-5.6 sol, and GPT-5.4 xhigh on composite score; the frontier rows retain an advantage on Adversarial cases, and GPT-5.4 xhigh on Temporal cases as well. The 27B regression is specific to the tested recipe, including its shorter training sequence length. The result does not show that small models generally replace frontier models. It shows that when an operational task is deliberately bounded, an open-weight model adapted to that setting can match a frontier API on its core executable requirements.

\bibliographystyle{tmlr}
\bibliography{references}
\clearpage
\appendix

\section{AI-assistance disclosure}\label{app:A}

Generative AI tools were used in preparing this work: for case authoring, for harness and analysis code, and for drafting text. The Figure~\ref{fig:1} artwork was produced with an image-generation model from an author-written specification; Figures 2 to 5 are plotted directly from the scored result files. The author reviewed and edited all generated content and takes full responsibility for the content of this paper.

\section{Reproduction}\label{app:B}

Source code repository: \url{https://github.com/continker/metrollm-bench} (tag \texttt{paper-v1.2}).

\subsection{Harness}\label{app:B.1}

\begin{itemize}
\tightlist
\item
  \textbf{Mock server}: FastAPI (port 8100), Pydantic-validated responses, per-case disruption state, six tools (\texttt{route\_planner}, \texttt{fare\_calculator}, \texttt{station\_info}, \texttt{disruption\_feed}, \texttt{line\_info}, \texttt{knowledge\_base}) and the terminal tool \texttt{submit\_assistant\_state}.
\item
  \textbf{Runner}: async \texttt{httpx}, OpenAI chat-completions interface, at most 20 tool rounds per case, 240 s timeout per LLM round, semaphore-gated concurrency (N=2 local, N=4 Mistral API, N=8 Azure).
\item
  \textbf{Cases}: \texttt{cases/\{system\}\_cases.json}, 156--167 cases per system across six systems, 955 total.
\item
  \textbf{Scorer}: deterministic Tier 1 (14 components); semantic-quality Tier 2 (8 components: 6 using Haiku and 2 scored programmatically), with language-model judgments cached per \texttt{(case,\ rubric)}.
\item
  \textbf{Framebook}: per-system YAML injected into the system prompt at runtime (terminology, fare rules, operating hours, cultural notes).
\end{itemize}

\subsection{Commands}\label{app:B.2}

The full command sequence, from partition build through teacher-trace generation, PEFT training, benchmarking, and the partition-filtered statistics, is maintained as \texttt{REPRODUCING.md} in the repository, where it stays in step with the code. The pipeline entry points are \texttt{scripts/build\_holdout\_split.py} and \texttt{scripts/slice\_cases\_by\_split.py} (partition), \texttt{harness/mock\_server.py}, \texttt{harness/runner.py}, and \texttt{harness/scorer.py} (evaluation), \texttt{harness/rule\_agent.py} (baseline), \texttt{scripts/peft/} (training-set construction, training, merge, GGUF export), and \texttt{scripts/score\_split.py}, \texttt{scripts/compute\_bootstrap\_heldout.py}, and \texttt{scripts/heldout\_percategory.py} (statistics).

\subsection{Hardware}\label{app:B.3}

All local inference ran on a single NVIDIA RTX 5090 with 32 GB of VRAM. The host used Ubuntu 24.04, a llama.cpp build from early April 2026, two parallel slots, flash attention, and bf16 compute. PEFT training used the same GPU through Hugging Face \texttt{transformers}, \texttt{peft}, and \texttt{bitsandbytes}.

\subsection{Model configurations}\label{app:B.4}

\begin{table}[!ht]
\centering
\small
\caption{Evaluation-time model configurations.}
\label{tab:5}
\vspace{2pt}
\begin{tabular}{@{}llll@{}}
\toprule
\textbf{Family} & \textbf{Temperature} & \textbf{Thinking} & \textbf{\texttt{reasoning\_effort}} \\
\midrule
Qwen 3.5 local & 0.0 & on & n/a \\
Gemma 4 local & 0.0 & n/a & n/a \\
Mistral API & 0.0 & n/a & n/a \\
GPT-5.6 Azure & 1.0 & n/a & medium (luna) / xhigh (sol) \\
GPT-5.4 Azure & 1.0 & n/a & medium / high / xhigh \\
\bottomrule
\end{tabular}
\end{table}

As Table~\ref{tab:5} shows, the OpenAI rows run at temperature 1.0, the only value GPT-5 deployments accept at the author's Azure subscription tier; GPT-5.4 full was run at all three effort levels and nano and mini at medium. That run-to-run variance is not separately measured. The local and Mistral evaluations use temperature 0.0, where Section~\ref{sec:3.4} measures a server-level spread of 0.23 Tier 1 points per system, and Qwen3.8-27B is the one local row served with sampling. Appendix~\ref{app:D} quantifies case-sampling uncertainty only; Section~\ref{sec:6} discusses the remaining sources as limitations.

GPT-5.6 was served through Azure OpenAI's Responses API: for this family the chat-completions endpoint rejects function tools combined with a reasoning-effort setting. The harness converts the Responses payload to the chat-completions shape before recording, so scoring is unchanged, and the run metadata records the dialect. Sol uses xhigh effort to match the GPT-5.4 full comparator row; luna uses the medium default, as nano and mini do. Both are single runs at temperature 1.0.

Rows added after v1 were served on llama.cpp build b10398 (the v1 rows on b8642) at the per-model configurations in Table~\ref{tab:6} (all with twenty tool rounds), which Section~\ref{sec:3.4} motivates. Re-benchmarking Qwen3.5-27B on b10398 under the otherwise unchanged v1 configuration gives 93.08 held-out Tier 1 against the published 92.32 on b8642: the serving build itself is part of the configuration, and cross-build comparisons should be read with that margin in mind.

\begin{table}[!ht]
\centering
\footnotesize
\caption{Per-model serving configurations for the rows marked \textsuperscript{3} in Table~\ref{tab:2} and for Llama 3.1-8B, all on llama.cpp b10398 with twenty tool rounds.}
\label{tab:6}
\vspace{2pt}
\begin{tabular}{@{}lp{4.4cm}rp{4.8cm}@{}}
\toprule
\textbf{Model} & \textbf{Decoding} & \textbf{Output budget} & \textbf{Notes} \\
\midrule
Muse Glimmer 30B & greedy (0.0) & 4096 & \texttt{reasoning\_strength: high} via chat-template kwargs \\
Qwen3.6-27B & greedy (0.0) & 16384 &  \\
Qwen3.8-27B & temp 1.0, top-p 0.95, top-k 20, min-p 0.0 & 16384 & vendor-recommended sampling; single run, seed 42 \\
GLM-4.7-Flash & greedy (0.0) & 4096 & larger budgets degrade (\S{}3.4) \\
Llama 3.1-8B & greedy (0.0) & 4096 & excluded from ranking (Appendix~\ref{app:F}) \\
\bottomrule
\end{tabular}
\end{table}

\subsection{Case generation}\label{app:B.5}

Case selection was manually curated as described in Section~\ref{sec:2.1}. \texttt{cases/generator.py} combines category-specific definitions with per-system metadata in \texttt{test\_pairs.json} and disruption templates in \texttt{events.yaml}. It calls \texttt{MetroGraph} and \texttt{FareCalculator} to derive routes and fares, then writes \texttt{(events,\ system\_context,\ ground\_truth)} records together with scoring weights and tolerances. Generation-time tests cover required fields, unique identifiers, station references, graph-valid routes, exact Category B fare consistency, outcome constraints, and category-specific structure. The committed \texttt{cases/\{system\}\_cases.json} files are the canonical evaluation artefacts used for all reported runs; reproducing the reported scores does not require regenerating them. Version 23 added fifteen cases selected after the documented scenario gap audit, all of which are pinned to the held-out partition.

\subsection{Apple Silicon hardware envelope}\label{app:B.6}

Exploratory measurements characterise local inference on two Apple Silicon systems, separate from the RTX 5090 configuration in Section~\ref{app:B.3}; they do not constitute an end-to-end kiosk latency study. The fanless 16 GB MacBook Air M2 sustains 39 tokens per second of 2B Q4\_K\_M decode after thermal stabilisation, the fan-cooled 96 GB M2 Max 108 tokens per second, and a 4B student is bandwidth-projected (not measured) to roughly 18 tokens per second on the Air. Throughput depends on architecture and quantisation rather than the adapter, and the 2B student used for the probe scores 85.8 Tier 1 on the fifteen-case probe set under deterministic Q4\_K\_M decoding. The figures describe decode throughput in the most favourable realistic operating state (lid open, on AC power), not user-perceived response time. The measurement protocol and thermal-curve scripts are documented in the repository's \texttt{REPRODUCING.md} and published as \texttt{continker/metrollm-bench-mac} on Hugging Face, so the measurements can be repeated on any Apple Silicon Mac.

\subsection{Independent ground-truth validation}\label{app:B.7}

A second annotator, independent of the project, audited the answer key itself, separately from the judge calibration in Appendix~\ref{app:E}. Fifty cases were sampled one per (system, category) cell (seed 42), covering all six systems and all eleven categories. A browser-based review interface presented the expected route and any closures on the network map, the system's fare rules beside the expected fare and breakdown, and the expected terminal kiosk state; the annotator gave a verdict on each of route, fare, and outcome, with a cannot-verify option throughout. The audit confirmed 38 of 38 expected routes, 36 of 38 fares, and 49 of 50 outcomes. One flag identified a confirmed answer-key error: in the passenger flip-flop scenario (H-006, present in all six systems), the expected fare was priced for two adults while the case's own acceptance pattern requires the one-adult total. The committed case files are corrected; no reported score was affected, because Category H does not weight \texttt{fare\_correct}. The two remaining flags raise design questions rather than defects and are discussed in Appendix~\ref{app:F}.

The follow-up investigation surfaced three further issues. Some disruption definitions list closure endpoints that are not adjacent in the graph, so the entry closes no edge; in all but two cases another closure in the same event still severs the route, and the two contradictory cases are retained unchanged and documented in Appendix~\ref{app:F}. The day-of-week label in 90 of 114 temporal contexts disagrees with the weekday of its timestamp; the open-or-closed determination is identical under either day in all 90 cases, so no expected outcome is affected. Finally, regeneration revealed that the generator's tie-breaking between lines that share track is not deterministic; the committed case files remain canonical (Section~\ref{app:B.5}), and the tie-break should be made deterministic before the case files are regenerated.

\section{PEFT training details}\label{app:C}

\subsection{Training-set composition}\label{app:C.1}

The two teacher models, Qwen 3.5 27B-dense and 35B-A3B MoE, run only on the 717-case training partition. We retain a trace when the teacher scores at least 90 percent on Tier 1, then deduplicate by \texttt{case\_id}, keeping the higher-scoring instance. The final set contains 540 traces from the dense 27B teacher and 60 from the 35B-A3B teacher. Its mean Tier 1 score is 99.0 percent. No held-out case enters a student's training data.

\begin{table}[!ht]
\centering
\small
\caption{Per-category PEFT training coverage (training examples / training-partition cases).}
\label{tab:7}
\vspace{2pt}
\begin{tabular}{@{}lrrr@{}}
\toprule
\textbf{Cat} & \textbf{In training} & \textbf{Train-partition cases} & \textbf{Coverage} \\
\midrule
A Routing & 91 & 91 & 100\% \\
B Fare & 73 & 73 & 100\% \\
C Disruption & 66 & 92 & 72\% \\
D Accessibility & 67 & 71 & 94\% \\
E Cultural & 27 & 28 & 96\% \\
F Policy & 72 & 72 & 100\% \\
G Multi-turn & 54 & 67 & 81\% \\
H Adversarial & 53 & 65 & 82\% \\
I Temporal & 33 & 68 & 49\% \\
J Hallucination & 48 & 71 & 68\% \\
K Compound & 16 & 19 & 84\% \\
\bottomrule
\end{tabular}
\end{table}

The Tier 1 threshold is a capability filter, not a category filter, so the coverage in Table~\ref{tab:7} reflects where the teachers succeed. Temporal reasoning (Category I) is undersampled because the base teachers score about 83 percent there and most traces fail the threshold. Compound Stress (Category K) has high proportional coverage but only nineteen cases in the training partition. It is therefore the category most vulnerable to memorisation, and the capacity-ceiling argument in Section~\ref{sec:4} does not depend on it.

\subsection{Hyperparameters}\label{app:C.2}

All four students follow the same recipe. The maximum sequence length is reduced from 4096 to 2048 at 27B to fit within 32 GB of VRAM.

\begin{itemize}
\tightlist
\item
  Base: Qwen 3.5 2B / 4B / 9B / 27B (bf16 weights)
\item
  Method: QLoRA with 4-bit NF4 base quantisation
\item
  LoRA rank 16, alpha 32, dropout 0.05
\item
  Target modules: \texttt{q\_proj,\ k\_proj,\ v\_proj,\ o\_proj,\ gate\_proj,\ up\_proj,\ down\_proj}
\item
  3 epochs, batch 2 with gradient accumulation 4 (effective batch 8)
\item
  Optimiser \texttt{paged\_adamw\_32bit}, lr 2e-4 cosine, 5 percent warmup
\item
  Seeds 42 (published as \texttt{continker/Qwen3.5-\{size\}-metro-v24}) and 43 (held internally) at every size, with seed 44 added at 2B, where seed sensitivity is largest, and evaluated in September 2026 under the same toolchain and serving configuration as the other seeds (llama.cpp b8642, 4096-token budget, greedy decoding); Section~\ref{sec:4} reports the per-seed comparison. The \texttt{-\/-seed} flag of \texttt{scripts/peft/train.py} propagates to LoRA initialisation, optimiser state, and the split between training and validation.
\end{itemize}

Re-running the published seed-42 artefact twice under that configuration reproduces its full-matrix Tier 1 within 0.5 points (76.61 and 75.74 against 76.24) while moving the held-out value by 2.1 and 0.3 points, and a second run of the seed-44 student differs from the first by 1.1 held-out points. Run-to-run variation for a 2B model is therefore about one held-out point, several times the 0.23 per-system spread measured for the 27B models in Section~\ref{sec:3.4}, and the 2B seed spread in Table~\ref{tab:4} should be read against that floor.

\begin{table}[!ht]
\centering
\small
\caption{PEFT training time and artefact sizes by student size.}
\label{tab:8}
\vspace{2pt}
\begin{tabular}{@{}lrrrr@{}}
\toprule
\textbf{Size} & \textbf{max\_seq} & \textbf{Training time / seed} & \textbf{Adapter} & \textbf{Q4\_K\_M GGUF} \\
\midrule
2B & 4096 & ~27 min & 42 MB & 1.2 GB \\
4B & 4096 & ~63 min & 100 MB & 2.6 GB \\
9B & 4096 & ~74 min & ~150 MB & 5.3 GB \\
27B & 2048 & ~103 min & 305 MB & 16 GB \\
\bottomrule
\end{tabular}
\end{table}

Table~\ref{tab:8} reports per-seed training time and final artefact sizes. Adapter merging, GGUF conversion, and Q4\_K\_M quantisation follow the scripted path in \texttt{scripts/peft/}, documented with exact library versions in the repository's \texttt{REPRODUCING.md}.

\subsection{Held-out partition}\label{app:C.3}

\texttt{scripts/build\_holdout\_split.py} creates the system-stratified 75/25 partition at seed 42. It assigns 717 cases to teacher-trace generation and student training, and reserves 238 for reporting. Fifteen gap-audit cases written after the first PEFT cycle are pinned to the held-out partition: \texttt{MARTA-D-016}, \texttt{MARTA-F-016}, \texttt{MARTA-F-017}, \texttt{BART-F-016}, \texttt{BART-B-016}, \texttt{CTA-F-016}, \texttt{CTA-C-018}, \texttt{DOHA-E-006}, \texttt{DOHA-E-007}, \texttt{DOHA-C-016}, \texttt{TRTC-B-016}, \texttt{TRTC-C-018}, \texttt{BJM-A-021}, \texttt{BJM-B-016}, and \texttt{BJM-C-022}. The per-system held-out fractions range from 24.7 to 25.2 percent. Appendix~\ref{app:G} describes the underlying artefacts.

\section{Statistical methodology}\label{app:D}

All non-PEFT scores come from a single run. Treating each case as a Bernoulli trial at 90 percent accuracy gives a standard error of $\sqrt{0.9 \cdot 0.1 / 955}$ $\approx$ 0.97 points for 955 cases and 1.9 points for the 238 held-out cases, or 95 percent half-widths of about 1.9 and 3.8 points. This is a conservative bound, since each case contributes a fractional component score rather than a binary outcome, so the observed intervals below are narrower; differences below one point remain within measurement noise on either partition.

\subsection{Bootstrap confidence intervals}\label{app:D.1}

The PEFT rows use the per-case mean over the two seeds, as in Table~\ref{tab:2}. The command \texttt{scripts/compute\_bootstrap\_heldout.py\ -\/-partition\ \{full,holdout\}\ -\/-n-resamples\ 5000} reproduces the intervals at seed 42. Table~\ref{tab:9} reports the results for the full 955-case matrix.

\begin{table}[!ht]
\centering
\small
\caption{Bootstrap confidence intervals for the full matrix.}
\label{tab:9}
\vspace{2pt}
\begin{tabular}{@{}lrrlrl@{}}
\toprule
\textbf{Model} & \textbf{n} & \textbf{T1 mean} & \textbf{T1 95\% CI} & \textbf{Comp. mean} & \textbf{Comp. 95\% CI} \\
\midrule
Qwen 27B base & 955 & 92.59 & [91.60, 93.51] & 90.69 & [89.83, 91.57] \\
Qwen 35B-A3B base & 955 & 92.22 & [91.23, 93.13] & 89.85 & [88.95, 90.69] \\
Qwen 27B + PEFT (n=2) & 955 & 91.50 & [90.49, 92.45] & 89.63 & [88.74, 90.49] \\
Qwen 9B + PEFT (n=2) & 955 & 91.26 & [90.29, 92.24] & 88.83 & [87.96, 89.63] \\
Qwen 4B + PEFT (n=2) & 955 & 90.94 & [89.94, 91.90] & 88.79 & [87.91, 89.68] \\
GPT-5.4 full xhigh & 955 & 90.93 & [90.05, 91.77] & 89.90 & [89.11, 90.65] \\
Qwen 4B base & 955 & 89.23 & [88.05, 90.35] & 87.00 & [85.94, 88.06] \\
\bottomrule
\end{tabular}
\end{table}

On the full matrix, the Tier 1 intervals have half-widths of about 1.0 point, roughly half the Bernoulli bound above because component scores are fractional. The corresponding intervals on the 238-case held-out partition are about twice as wide. This loss of precision explains why the individual held-out differences do not reach significance.

\subsection{Paired bootstrap}\label{app:D.2}

The paired bootstrap resamples cases while preserving the pairing between models. An interval that excludes zero is significant at $\alpha$ = 0.05. Table~\ref{tab:10} reports the four principal comparisons for both partitions.

\begin{table}[!ht]
\centering
\small
\caption{Paired-bootstrap Tier 1 comparisons on the held-out and full partitions.}
\label{tab:10}
\vspace{2pt}
\begin{tabular}{@{}llll@{}}
\toprule
\textbf{A} & \textbf{B} & \textbf{$\Delta$T1 held-out (n=238)} & \textbf{$\Delta$T1 full (n=955)} \\
\midrule
4B + PEFT & GPT-5.4 full xhigh & $-$0.05 [$-$1.80, +1.70] & +0.01 [$-$0.93, +0.94] \\
4B + PEFT & Qwen 4B base & +1.97 [$-$0.17, +4.16] & +1.72 [+0.72, +2.74] \\
27B + PEFT & Qwen 27B base & $-$0.94 [$-$2.26, +0.39] & $-$1.09 [$-$1.82, $-$0.38] \\
9B + PEFT & Qwen 35B-A3B base & $-$1.10 [$-$3.34, +1.07] & $-$0.95 [$-$1.91, +0.01] \\
\bottomrule
\end{tabular}
\end{table}

The full-matrix composite differences also exclude zero: $-$1.11 {[}$-$1.95, $-$0.31{]} for the 4B student against GPT-5.4, $-$1.06 {[}$-$1.72, $-$0.40{]} for the 27B regression, and +1.79 {[}+0.90, +2.71{]} for the 4B student against its base. Every held-out interval includes zero because 238 cases do not provide enough power to establish an individual pairwise difference. On the full matrix, however, the 4B PEFT gain and 27B regression are significant on both Tier 1 and composite score. The 4B student ties GPT-5.4 full xhigh on Tier 1 (+0.01 {[}$-$0.93, +0.94{]}) while trailing by approximately one composite point. Section~\ref{sec:4} bases the capacity-ceiling conclusion on these intervals and on the consistency of the trend across partitions and seeds.

\subsection{What the bootstrap does not show}\label{app:D.3}

\begin{itemize}
\tightlist
\item
  The bootstrap captures case-sampling variation only. Section~\ref{sec:4} reports training-seed variation separately from two seeds (three at 2B), ranging from a 3.96-point spread at 2B to 0.09 at 27B on the full matrix. The two analyses answer different questions and are not combined.
\item
  The per-category differences in Figure~\ref{fig:3} are not paired-tested. Held-out category counts range from 11 to 32, so some apparent differences are within sampling noise.
\item
  The four principal comparisons were selected by the author. A confirmatory study should apply a correction for multiple comparisons.
\end{itemize}

\section{Scoring-stack calibration and judge rubrics}\label{app:E}

Claude Haiku 4.5 (\texttt{claude-haiku-4-5-20251001}), Anthropic's small, low-cost model, provides the language-model judgments for six of the eight Tier 2 components, sometimes alongside structural checks. It is affordable enough to score every applicable judge rubric throughout the evaluation, and its refusal behaviour is well characterised. All judgments are cached. No Anthropic model is evaluated as an agent, which also avoids comparing an Anthropic agent with an Anthropic judge; the author has no affiliation with Anthropic, which acts solely as an inference provider for the judge.

For calibration, two annotators independently rated the same 100 case-rubric pairs, drawn from 100 different cases across all six systems and ten of the eleven categories; the rated responses are GPT-5-mini outputs from a single benchmark run. The author rated first, with each automated score revealed only after the rating was locked; the second annotator, independent of the project, later rated the identical pairs in a session that displayed no judge output at any point. The sample is stratified across six Haiku-using Tier 2 rubrics and the deterministic Tier 1 \texttt{cultural\_accuracy} check used by the deployed scorer, with fourteen or fifteen pairs per rubric, and is enriched for non-full-credit automated outputs. Deployed scores are mapped to a common ordinal 0/1/2 scale. Author versus judge: 82 percent exact agreement, 97 percent within one point, $\kappa_w$ = 0.53 (95\% bootstrap CI {[}0.27, 0.75{]}), moderate under \citet{landis1977}; on the 85 Haiku-scored pairs, 84 percent exact agreement and $\kappa_w$ = 0.56. Second annotator versus judge: 69 percent exact, 93 percent within one point, $\kappa_w$ = 0.02 (CI {[}$-$0.12, 0.18{]}). Between the annotators: 76 percent exact, 97 percent within one point, $\kappa_w$ = 0.25 (CI {[}0.02, 0.46{]}). Under prevalence-robust Gwet AC2 \citep{gwet2008} the three pairings score 0.89, 0.81, and 0.89; the $\kappa_w$ spread reflects the score marginals, since the second annotator awarded 2/2 on 89 of 100 pairs. The same leniency produced two clear misses: on CTA-H-003 (\texttt{safety\_response\_quality}) and DOHA-D-006 (\texttt{scope\_adherence}) the second annotator awarded full credit where the author and the judge both scored zero.

Two effects explain the difference between the high exact agreement and moderate $\kappa_w$. First, the H-014 scenic-route cases for TRTC, MARTA, and CTA expose a definitional disagreement over \texttt{safety\_response\_quality}. The judge penalises an offer that violates a system-prompt constraint, while both annotators credit the valid route to the destination. Second, four rubrics are highly imbalanced: 13 to 15 of the 14 or 15 cases receive 2/2. The resulting per-rubric $\kappa$ approaches zero despite exact agreement between 73 and 93 percent. Across the more informative rubrics, $\kappa_w$ is 0.20 for \texttt{safety\_response\_quality}, 0.68 for \texttt{scope\_adherence}, and 0.69 for \texttt{policy\_acknowledged}.

Three rubric changes were made during development. We moved \texttt{cultural\_accuracy} to a deterministic keyword check after the judge introduced unsupported elaboration requirements. We added ``do not penalise'' clauses and a 60-minute last-train threshold to the temporal and safety rubrics. We also supplied the rejection reason as context for adversarial cases. The author stopped tuning at that point to avoid overfitting the calibration set. Calibration against a judge from another model family remains future work. No headline claim is based on Tier 2 alone: the central deployment comparison uses deterministic Tier 1, while the composite score is a secondary measure that necessarily includes Tier 2.

\needspace{6\baselineskip}
Representative abridged rubric text (full prompts in \texttt{harness/judge.py}):

\begin{itemize}
\tightlist
\item
  \texttt{advisory\_content\_correct} (5 pt): ``Given the disruption context and the expected advisory keywords, does the model's advisory banner substantively cover the required points? Score 5 for full coverage, 3 for partial with a key omission, 0 for missing or hallucinated content.''
\item
  \texttt{scope\_adherence} (5 pt): ``Did the assistant stay within transit-kiosk scope? Penalise offers of taxi, ride-share, or out-of-scope services; do not penalise informational mentions that are declined or redirected.''
\end{itemize}

\section{Limitations and threats to validity}\label{app:F}

\textbf{What the benchmark measures.} MetroLLM-Bench tests whether a model can parse a transit scenario, call six deterministic tools with the correct arguments, and synthesise a structurally valid terminal state. The state includes an outcome, fare breakdown, kiosk action, passenger-facing message, and any required advisory banners. Tier 2 additionally measures semantic quality through six components that use a language-model judge and two programmatic checks.

\textbf{What the benchmark does not measure.} The evaluation does not cover end-user satisfaction, payment or ADA-device integration, multi-session continuity, PII handling, calibrated refusal confidence, or behaviour under strict latency targets. All responses are in English, including those for non-English systems. Category H tests a defined set of transit-domain attacks rather than general adversarial robustness.

\textbf{Scope of the claims.} The results do not imply that language models replace programmers; tool implementation, payment integration, and harness maintenance remain code-level work. They do not establish PEFT as universally better than full fine-tuning, prompt engineering, or MoE routing. Nor do they show that very small models replace large ones: performance falls sharply below the 4B Qwen base.

\textbf{Known answer-key defects retained.} Independent validation (Appendix~\ref{app:B.7}) left four issues documented rather than fixed, because fixing them would alter the fixed evaluation set mid-campaign. In two disruption cases (TRTC-C-007, CTA-C-008) the maintenance closure lists only non-adjacent endpoints, so no edge is actually closed and the expected route legitimately traverses the nominally closed stretch; correcting them would flip the expected outcome, with an effect of at most 0.3 composite points, below the measured seed-to-seed variation. One MARTA cultural case (MARTA-E-003) expects the eating-and-drinking rule to be stated although neither the MARTA framebook nor its knowledge base contains it, so the case rewards world knowledge rather than grounding; adding the missing framebook note would change the system prompt for every MARTA case and is deferred to the next full evaluation round. And one accessibility case pairs a stated mobility impairment in prose with an adult-fare expectation; whether a kiosk should infer concession eligibility from prose is an open design question the benchmark does not test. One multi-turn case (BJM-G-009) states an elevator requirement in the conversation that its expected terminal outcome does not carry; whether the outcome should encode that requirement is likewise a design question rather than a defect in the expected route or fare.

\textbf{Excluded models.} Gemma 4 E2B and E4B are excluded because 28 to 33 percent of their cases exhaust the twenty-round budget without a valid \texttt{submit\_assistant\_state}. Their floor-effect zeros are not comparable with models that terminate normally on at least 98 percent of cases. The raw composite scores are 42.8 for E2B and 66.8 for E4B. Both variants nevertheless achieve 99.9 percent scope adherence and 100 percent no-tool-hallucination, identifying structured-output synthesis as the failure rather than tool selection. Gemma 4 26B-A4B terminates normally and remains ranked. We did not raise the round limit during the evaluation because doing so would have doubled Gemma wall-clock time; a deployed kiosk would also face a tighter latency budget, not a looser one. Llama 3.1-8B is excluded for the same reason at greater severity: 74.5 percent of its cases end in a recorded harness error and 36.6 percent never call \texttt{submit\_assistant\_state} at all, for raw held-out scores of 38.70 composite and 39.52 Tier 1. As with Gemma E2B and E4B, this is a structured-output and protocol failure rather than a capability ranking; it is also a dated model, evaluated as the most recent open dense Llama in its size class.

\textbf{Model selection.} The matrix was assembled incrementally from models available on the local RTX 5090 and through APIs already accessible to the author. Claude is the most notable omitted family. Because Claude Haiku supplies the language-model judgments used in Tier 2, excluding Claude agents avoids a same-family agent-judge comparison. Llama 3.1-8B is excluded from the ranking as described above.

\textbf{Pretraining familiarity.} MARTA, BART, and CTA are well-documented US networks, whereas Doha, Taipei, and Beijing are less prominent in English-language sources. Models could therefore benefit from unequal pretraining familiarity. The held-out per-system results do not show such an advantage. Across the seven models of Figure~\ref{fig:3}, the mean difference between US and non-US Tier 1 scores ranges from $-$2.6 to +0.5 points. Four models score higher on the non-US systems by more than a point, the 4B and 27B students are level to within 0.05 points, and GPT-5.4 full xhigh favours the US systems by 0.5 points.

Beijing and Taipei are among the highest-scoring systems despite being the largest and less well documented operationally. Qwen 27B base reaches 94.8 and 94.3 Tier 1 on them. The lowest cells for that model comparison occur on BART (88.8 for Qwen 27B base) and CTA (87.0 for GPT-5.4 xhigh). These results are consistent with the in-context design, in which the framebook and tools provide every fact needed by a case, MARTA-E-003 excepted. Per-system held-out samples contain only about forty cases, however, so individual cells remain noisy.

\textbf{Teacher-model dependence.} The PEFT students distil from Qwen 3.5 27B-dense and 35B-A3B, both of which appear in Table~\ref{tab:2}. The Qwen rows are therefore same-family comparisons, not independent baselines. The held-out partition prevents direct case leakage because all teacher traces come from the training partition. No student exceeds either teacher on held-out cases: the 4B student scores 91.3 on Tier 1, compared with 92.3 for the 27B teacher. The result is compression to teacher-class quality, not improvement over the teacher.

\textbf{Additional sources of uncertainty.} The Haiku judge is pinned to \texttt{claude-haiku-4-5-20251001}. If the provider silently changes the weights behind that identifier, fresh judgments may differ from the cached decisions used here. The framebooks and cases were written by one author with AI assistance; another authoring team, particularly one involving a transit operator, might produce different ground truth. That difference could favour models whose training data more closely resembles the present author's prose. Finally, the harness gained a batched \texttt{line\_info} form and a multi-disruption fix during the evaluation cycle. Those changes affect wall-clock and token-cost reporting, but not Tier 1 correctness, which does not depend on tool count.

\textbf{Contamination.} The case files have been public since 19 June 2026. Two ranked models were released after that date without a documented training cutoff, Muse Glimmer 30B (10 August 2026) and Qwen3.8-27B (14 August 2026), so exposure of their training data to the public cases cannot be excluded from the available documentation; GPT-5.6's documented cutoff (February 2026) precedes publication, and every other model in Table~\ref{tab:2} was released before the cases were public.

\section{Data and model artefacts}\label{app:G}

The public \href{https://github.com/continker/metrollm-bench}{MetroLLM-Bench repository} carries the benchmark itself: the six system datasets and framebooks, the 955 committed case files with pre-sliced train and held-out variants, the partition specification (\texttt{data/splits/v23\_holdout75\_seed42.json}), the complete harness (mock server, runner, scorer, judge, rule-based agent), the PEFT pipeline scripts, and the step-by-step reproduction guide (\texttt{REPRODUCING.md}). The four PEFT students are published on Hugging Face as \texttt{continker/Qwen3.5-\{2B,4B,9B,27B\}-metro-v24}, each with the LoRA adapter, the Q4\_K\_M GGUF, and a model card; the Apple Silicon measurement package is \texttt{continker/metrollm-bench-mac}. The per-case scored result files and judge caches underlying every number in this paper, and the 600-trace training set, are archived by the author and are regenerable end to end from the repository. Every cell in Table~\ref{tab:2} and Figures~\ref{fig:2}--\ref{fig:5} is traceable to a \texttt{(system,\ model\_tag,\ case\_id)} triple via those scored files.

\end{document}